\documentclass[11pt]{article}
\usepackage{ranlp97}
\usepackage{epsf}

\title{Segmentation of Expository Texts by Hierarchical Agglomerative
Clustering}
\author{Yaakov Yaari\thanks{\ \ This work is part of PhD thesis done in the
faculty of Mathematics and Computer Science, Bar Ilan University,
under the supervision of Prof. Yaacov Choueka}\\
Bar-Ilan University\\
Ramat-Gan 52900, Israel\\
\texttt{yyaari@macs.biu.ac.il}}

\begin{document}

\maketitle

\begin{abstract}
We propose a method for segmentation of expository texts based
on hierarchical agglomerative clustering. The method uses paragraphs
as the basic segments for identifying hierarchical discourse structure
in the text, applying lexical similarity between them as the proximity
test. Linear segmentation can be induced from the identified structure
through application of two simple rules. However the hierarchy can be
used also for intelligent exploration of the text. The proposed
segmentation algorithm is evaluated against an accepted linear
segmentation method and shows comparable results.
\end{abstract}

\section*{Introduction}
The interest in expository texts comes, among others, from their
widespread use as online information resources. Information retrieval
gives us methods for scoring document collections based on their
relevance to a query. However once we are about to browse a document,
or extract some specific information from it, an interest arises for
deeper analysis of the text.

The kind and depth of the analysis depends on the reader. In the case of
text extraction, the ``reader'' is the text understanding system,
which implies a need for rather deep semantic analysis
~\cite{iwanska91,soderland94,hahn90}. However for (human)
browsing and reading in free expository texts, we can suffice with
delineating the structure of the text and provide easy access to the
discovered substructures. This kind of discourse segmentation is thus
a critical task for exploratory reading of the retrieved text.

This article presents an approach for discovering discourse structure in free
expository texts. The identified structure can be used in various tasks such as
text browsing and summarization. Section~\ref{discSeg} surveys other
approaches to discourse segmentation, in particular those based on
lexical cohesion metrics. Section~\ref{bottomUp} details
the proposed method for identifying a hierarchical discourse structure in the
text, based on the hierarchical agglomerative
clustering method, while using common lexical cohesion metrics. We show how,
through application two simple rules, linear discourse
segmentation can be recognized in the discovered structure. The output of the
algorithm is analyzed and evaluated in section~\ref{eval}. Finally
section~\ref{conclusion} concludes the paper and outlines future work.

\section{Discourse Segmentation Methods}
\label{discSeg}
Two main approaches can be seen in discourse segmentation of
free text: the multiple-source approach where multiple kinds of
evidence are used to determine discourse boundaries and relations, and
the lexical cohesion approach, where lexical cohesion (or lexical
similarity) is the sole criteria for boundary detection. These two
approaches are discussed in the following subsections.

\subsection{Multiple Source Methods}
\label{multPart}
With the dominant discourse analysis theories of today \cite{grosz86,mann87,grosz95}
there is no simple computational way to determine the detailed
discourse structure from the free text. Detailed
structure may include participants intentions, coherent discourse
segments, their functions and inter-relations.

To reach this level of understanding, researchers are forced to use
multiple sources of evidence and apply it in some adaptive manner. A
good example is in~\cite{litman94} which uses prosodic features, cue
phrases, and NP references, and applies machine learning methods for
the analysis of verbal discourse. Another
approach~\cite{kurohashi94,miike94} is to identify inter-sentential
discourse relationships using sources of evidence like cue phrases,
topic words/phrases, and grammatical and lexical similarity between
sentences. A complex decision rule sets was developed to map cue word
patterns to potential discourse functions or relationships.

\subsection{Lexical Cohesion Methods}
\label{lexCohesion}
When the source text is truly free text, methods that attempt to
identify detailed discourse structure tend to be brittle. Researchers
then turn to simpler but reliable evidence, in particular, lexical
cohesion.

Lexical cohesion between two discourse segments is an indicator of
textual coherence, and is achieved when the segments contain words
which are similar or semantically related ~\cite{halliday76}. We will
say that a discourse boundary exists between the two segments if the
lexical cohesion between them, as computed by some similarity metrics,
for example using the cosine distance between the term
vectors~\cite{salton89}, falls below some threshold.

Two related methods~\cite{kozima93,hearst94} use a concept of text
window, within which they compute a lexical cohesion function. By
moving the window $W$ over the text, they form a linear plot of the
lexical cohesion as a function of the word position, $w_i$ (where the
window is centered). A discourse boundary is assigned to $w_i$ if that
value falls below a threshold. Kozima uses a semantic network with
words at the nodes and edges indicating their semantic relation as
computed from a MRD. The lexical cohesion function computes $w_i$ by
spreading activation on the semantic network for each word in the
window $W$ and summing the output at $w_i$.  Here, then, the lexical
cohesion takes into account the similarity of the words based on their
definition in a dictionary. Hearst splits the window $W$ to two
halves, the one to the left of $w_i$, and the one to its right, and
determines the term vector of each. Term vectors consist simply the
counts of each open class words in the window. She then computes the
lexical cohesion function at $w_i$ by evaluating the similarity
between the two terms vector using the cosine distance formula.

While the above methods produce linear segmentation, some attempts
have been made to identify a more elaborated structure. The lexical
chaining method~\cite{morris91} attempts to determine the hierarchical
intention structure \cite{grosz86} by identifying lexical chains that
run through the text. The lexical cohesion between words in a chain is
determined using various relations defined over the Roget thesaurus,
however the algorithm is only implemented manually. A more practical
approach is to build, from the text, a graph with paragraphs at the
nodes, and their lexical similarity at the edges~\cite{salton94}. By
setting a threshold we can than identify strongly-connected subgraphs
which correspond to inter-related paragraphs. This structure can be
used both to improve text retrieval and for identifying themes for
text browsing.

\section{Segmentation By Hierarchical Agglomerative Clustering}
\label{bottomUp}
The proposed segmentation process consists of three main phases:
\begin{enumerate}
\item Morphological analysis.
\item Hierarchical agglomerative clustering of text segments.
\item Boundary detection.
\end{enumerate}

\subsection{Morphological Analysis}
\label{token}
The purpose of this phase is to determine the terms to be used as
content words in the following phase. The phase consists of the
following steps:

\begin{enumerate}

\item Tokenization. Convert the raw text, through regular
expression recognizer, to streams of tokens: words, numbers and
special symbols. 

\item Perform part-of-speech tagging~\cite{brill94}. This step filters
open class words, adjectives, verbs, adverbs, and nouns, to the next step.

\item Determine the general significance of each word $i$,
$Gsig_i$. In this experiment we use IDF as the measure for general
significance, using frequency-in-files information from the BNC corpus
\cite{leech92}:
\begin{equation}
\label{sig}
Gsig_i=IDF_i=\log{\frac{N}{N_i}}
\end{equation}
where $N$ is the number of files in the corpus and $N_i$ is the number
of files containing word $i$.

\item Stemming. Replace each word by its stem, $r_i$ (Porter's
algorithm is used here~\cite{porter80}). The general significance
$Gsig_i$, associated with $r_i$, is the minimum $Gsig$  over all words $j$
having this stem: $Gsig_i=\min_{j,r_j=r_i}{Gsig_j}$. This has the effect
of counting all instances of a given stem as a single concept.
\end{enumerate}

\subsection{Hierarchical agglomerative clustering of text segments}
\label{algorithm}
The main motivation behind the proposed algorithm is discovering a
structure in text. The bottom-up Hierarchical Agglomerative Clustering (HAC)
algorithm is a widely used clustering method in information retrieval
\cite{everitt80}, psychology \cite{milligan80}, linguistics
\cite{kessler94}, and elsewhere. 

When applying hierarchical agglomerative clustering on text segments
the algorithm successively grows areas of coherence at the most
appropriate place, thus forming a text structure. A similar
approach~\cite{maarek94} uses HAC to determine a hierarchical
bookshelf from a given set of documents.

The HAC algorithm for discourse segmentation, based on paragraphs as
the elementary segments, is shown in Figure~\ref{alg}.
\begin{figure}[h]
\begin{description}
\item {\bf Partition} the text to elementary segments (=paragraphs). 
\item {\bf While} more than one segment left {\bf do}
\begin{description}
\item {\bf Apply} a proximity test to find the two most similar
consecutive segments, $s_i$, $s_{i+1}$.
\item {\bf Merge} $s_i$, $s_{i+1}$ into one segment.
\end{description}
\item {\bf End while}
\end{description}
\caption{Hierarchical agglomerative clustering of text segments}
\label{alg}
\end{figure}

\begin{figure*}
\center {
\epsfxsize=6in
\leavevmode
\epsfbox{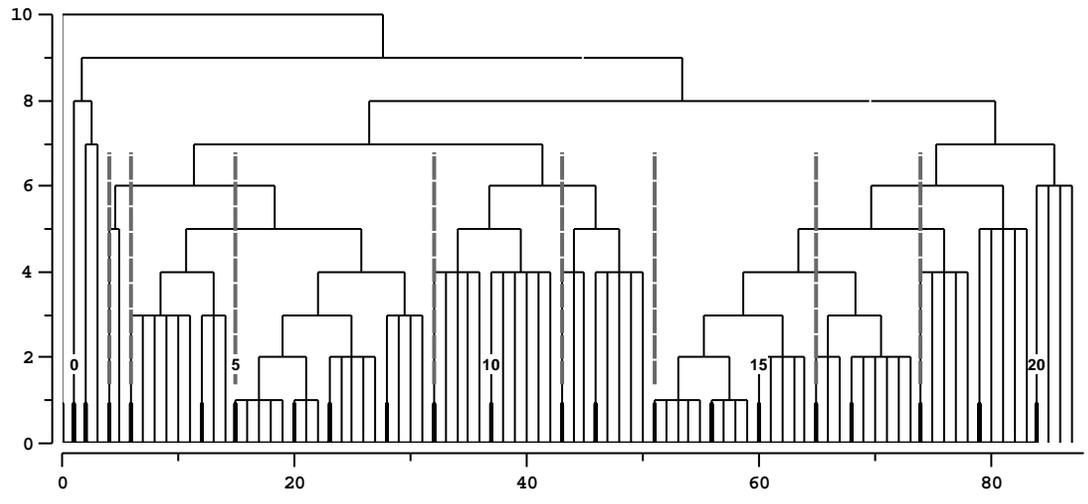}
\caption{\label{starDendro} Paragraph dendrogram of the
\emph{Stargazers} article. The leaves in the dendrogram are paragraphs
shown as a sequence of equal-length vertical lines - the
paragraph's sentences. The scale below the X axis shows sentence
numbers and the one above paragraph numbers (placed at end of the
respective paragraphs). Gray dashed vertical lines show the computed
boundaries.}  }
\end{figure*}

\begin{figure*} 
\center {
\epsfxsize=6in
\epsfysize=2.4in
\leavevmode
\epsfbox{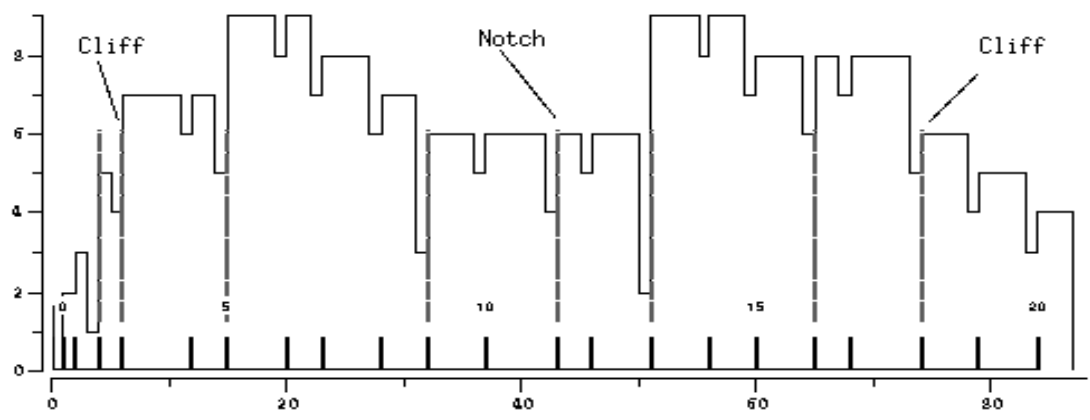}
\caption{\label{starOut} Outline of the \emph{Stargazers} article. The
graph plots the depth of each paragraph in the dendrogram, i.e., its
path length to the dendrogram root. The notches indicate the depth of the merge points. Gray dashed vertical lines are
segment boundaries. Paragraphs marks are shown
above the X scale, sentence marks below it}
}
\end{figure*}

Figure~\ref{starDendro} shows the result of the algorithm for the
\emph{Stargazers} text~\cite{hearst94}, in a dendrogram
representation. Figure~\ref{starOut} shows the
corresponding outline representation, which plots the depth of the
nesting of the paragraphs in the dendrogram, that is, the path length
from the paragraph node to the dendrogram root. The gray dashed vertical
lines both plots indicate the segment boundaries. Determination of
these boundaries is discussed in the next section. 

The algorithm successively grows ``coherent'' segments by
appending lexically related paragraphs, or by merging larger
segments. The result is hierarchical structure , called
\emph{dendrogram}, where text segments correspond to its subtrees. We
propose that the dendrogram represents the internal hierarchy
of the text discourse, similar to an intention structure~\cite{grosz86}.

Using paragraphs as the elementary segments for the
algorithm makes sense for a number of reasons. The paragraph is a
universal linguistic structure, representing a coherent textual
segment \cite{chafe79,longacre79,kieras82}. Allowing a boundary in the
middle of the paragraph is thus counter to
the author's intention. In addition, the size of a paragraph, unlike a
sentence, contains sufficient lexical information for the proximity
test.

Note that unlike general HAC applications, where at each stage we
compute the proximity of the newly merged object to all other available
segments, in our case we compute only the proximity of the segment to its two
neighbors. This is because we require that the linear order in the text
will be preserved in the structure. The implication on complexity is
that while general HAC algorithm takes an order of $O(N^2)$ steps, ours takes
only $O(N)$.

The proximity test selects the closest pair of segments. The test is
based on repetition of words, a well-recognized indicator for
lexical cohesion (see \cite{hearst94} for more references). The test
computes the cosine between the representative term vectors of
the segments \cite{salton88}:
\begin{equation}
Proximity(s_i,s_{i+1})=\sum_{k=1}^n\frac{w_{k,i}\cdot w_{k,{i+1}}}{||s_i||\cdot||s_{i+1}||}
\end{equation}
Where $s_i$ is the term vector representing segment $i$, $||s_i||$ is
its length, $\sqrt{\sum_{k=1}^n{w_{k,i}^2}}$, and $w_{k,i}$ is the weight of word $k$ in segment
$i$.

\begin{equation}
\label{wki}
w_{k,i}=f_{k,i}\cdot\frac{f_i}{f_{max}}\cdot Gsig_i
\end{equation}
The word weight $w_{k,i}$ is the product of three factors -  $f_{k,i}$, the
frequency of the word in the segment, serves as the in-segment
factor, $\frac{f_i}{f_{max}}$, the relative frequency of word $i$ in
the text, is an in-text factor, and $Gsig_i$ is the general word significance~(\ref{sig}).

\subsection{Boundary Detection}
\label{boundary}

The algorithm for boundary detection in the dendrogram makes use of
size and depth attributes of a segment. As indicated above, a segment
corresponds to a subtree in the resulting dendrogram tree. The segment
size is defined as the number of leaves, i.e. paragraphs, it contains. Its depth
is defined as the longest path in the subtree from the root to the
leaves. Thus, a size 1 segment is a single paragraph.

With these definitions, the algorithm for boundary detection is
stated in Figure~\ref{boundDetect}.
\begin{figure}[h]
\begin{description}
\item {\bf For each} node in the dendrogram tree $T$ {\bf do}
\begin{description}
\item {\bf Let} $S_1$ and $S_2$ be the two segments being merged at the node, such that
$size(S_1) \ge size(S_2)$. 
\item {\bf Set} a boundary between the two segments if one of
the following two rules holds:
\begin{description}
\item {\bf The notch rule:} \\$size(S_1) > n \land size(S_2) > n$
\item {\bf The cliff rule:} \\$size(S_1) > n \, \land \, size(S_2) \le n \,  \land \\
depth(S_1)-depth(S_2) > m$
\end{description}
\end{description}
\item {\bf End for each}
\end{description}
\caption{Algorithm for identifying boundaries in a dendrogram}
\label{boundDetect}
\end{figure}

The algorithm defines two rules to identify boundaries. The notch rule
constrains segments across boundaries to be of a significant size. We found that $n=1$ gives
maximum boundary information without adding false boundaries, that is,
it allows a paragraph that is cohesive with its neighbor
to be merged with it without creating a boundary between them.
The cliff rule relaxes
the notch rule, allowing one of the segments to be smaller than $n$ if
the difference between their depths is larger than a threshold
$m$. Such boundaries indicate remotely related segments and are seen
as high cliffs in the outline plot. The minimum for $m$ was set
experimentally to $depth(T)/5$.

Cases of the notch rule are seen, as the name implies, as deep notches
(deeper than 1 $(=n)$, in our case) in the outline view. See, for example, Figure~\ref{starOut},
between paragraphs 11 and 12. Cases of the
cliff rule may indicate a setting (or introduction) segment at the beginning of a
larger text segment, or a summary segment at its end. These setting and
summary segments consist of paragraphs, each discussing a different
topic. This creates a build-up effect (in case of setting) or a
fall-off effect (in case of summary). Cliff boundaries happen less frequently
than notches. For example, in Figure~\ref{starOut} they
appear between paragraph 3 and 4, and between paragraphs 18 and
19. In the last case, paragraphs 19, 20 and 21 can be regarded as a conclusion
section for the whole article. While the bulk of the article talks
about the special case of the earth and the moon, and their life-enabling
conditions, the last paragraphs summarize the conditions for life
existence in a solar
system and future research directions to be undertaken by astronomers.

\section{Evaluation}
\label{eval}

We have used the \emph{Stargazers} article, discussed in the previous section, as the test bench for
evaluation. \emph{Stargazers} is an expository text that discusses the conditions
fore evolution of life in solar systems. What makes it
particularly useful is that segmentation data is available
both as produced by Hearst's TextTiling algorithm,
which is robust and gives good results, and as produced by human
judges~\cite{hearst94}.

Comparing the results of TextTiling and Hierarchical Agglomerative
Clustering for boundary detection shows impressive
matching. Table~\ref{compHearstHuman} compares the results of the two
algorithms against those of the human judges. The boundaries for the
human judges are those with agreement of 3 or more among the 7 judges,
and are considered the correct boundary set. The P and R columns give
the precision and recall relative to that correct set.
{\scriptsize
\begin{table}[h]
\begin{center}
\begin{tabular}{| l | l | r | r |}\hline
\multicolumn{1}{| c }{} & 
\multicolumn{1}{| c }{Boundaries} &
\multicolumn{1}{| c }{P} &
\multicolumn{1}{| c |}{R}\\\hline
Human judges\footnote{Paragraphs with 3 out of 7 agreement} & 2 3 5 8 9 12 13 16 18 & 100 & 100\\\hline
TextTiling & 3 5 9 11 13 16 18 20 & 69 & 56\\\hline
HAC & 2 3 5 9 11 13 16 18 & 87 & 78\\\hline
\end{tabular}
\end{center}
\caption{\label{compHearstHuman} Performance of discourse segmentation
algorithms}
\end{table}
}
These results, while not yet very extensive, are encouraging. The
reason for the good match between boundaries determined by the two
algorithms is that in both cases boundaries are set when they separate
segments of low lexical cohesion. The main difference is the way these
segments are determined - fixed size in case of TextTiling, versus
variable size in case of HAC. 

\begin{figure*}
\center {
\epsfxsize=6in
\epsfysize=1.5in
\leavevmode
\epsfbox{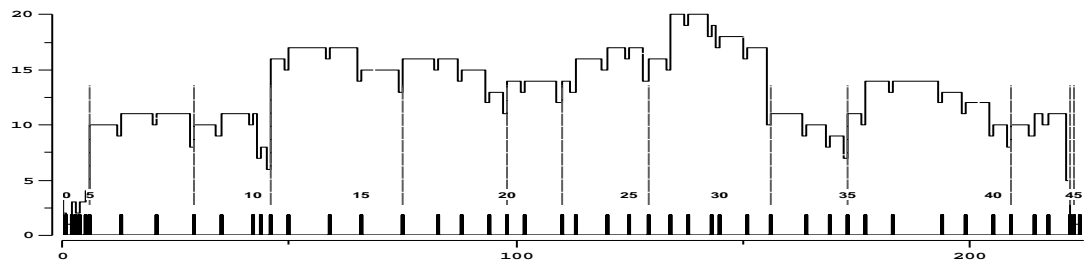}
\caption{\label{desertOut} Outline of ``How to Make a Desert'', Discover Magazine,
2/96}
}
\end{figure*}

But the HAC algorithm provides richer information than just linear
segmentation. The hierarchical clustering created by the algorithm
identifies a nested outline of the text. For example, we can deduce from the outline of
Figure~\ref{starOut} that
the \{17...18\} segment is part of a larger \{14...18\}
segment. Indeed, the enclosing segment is about binary/trinary systems
while the subsegment \{17...18\} is about their low
probability. Similarly we can deduce that the segment
\{12,13\} is more lexically-related to \{10,11\} than to
\{14,15,16\}. Another example is the typical build-up of a setting and
fall-off of a summary, seen in coherent texts (see Figs.~\ref{starOut} and
~\ref{desertOut}). This information may
help us later in constructing a table-of-content visual representation
of the text.

\begin{figure*}
 \center {
\epsfxsize=6in
\epsfysize=1.5in
\leavevmode
\epsfbox{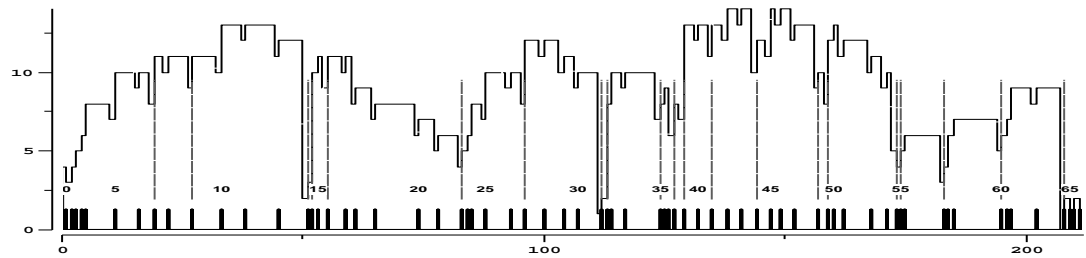}
\caption{\label{geneOut} Outline of ``Special issue: The Year in Science --
Genetics'', Discover Magazine, 1/96}
}
\end{figure*}

Figure~\ref{geneOut} presents an outline of a case of
``non-coherent'' text. The article is not about any specific subject but
rather a survey of special events in genetic engineering during
1995. The outline shows deep notches, following paragraphs 13, 22, 31,
35, and 49. These are the exact boundaries between the main articles
in the text. Unlike the former examples, here there is  no
fall-off summary at the end. This is expected since the ending
paragraphs are, in fact, a series of three tiny independent
articles, following paragraphs 55, 57 and 60. 

\section{Conclusions and Future Work}
\label{conclusion}
The main topic for research in the HAC algorithm is the proximity
test. At the moment it is a rather simple lexical similarity test, so
some modifications are possible in the way words are weighted (see
Equations~\ref{sig} and \ref{wki}). A more radical approach is using
concept vectors like in WordSpace \cite{schutze93}. Other sources of
information can be used to complement lexical similarity. In
particular, evidence involving cue phrases and part-of-speech patterns
can be processed, using previously-trained decision trees, to augment
the lexical similarity function (\cite{litman94}).

Another research direction is table-of-content production. The
clustering produced by the HAC algorithm provides the necessary
structure information. The main task here, and a major research topic, is
identification of topics, or titles, for the segments.

Finally, while the comparison with the TextTiling algorithm and the human
judges is promising, a methodical
evaluation of additional texts is required.

\bibliographystyle{ranlp}
\begin{scriptsize}
\bibliography{texplore}
\end{scriptsize}

\end{document}